\documentclass[conference]{IEEEtran}
\IEEEoverridecommandlockouts
\usepackage{cite}
\usepackage{amsmath,amssymb,amsfonts}
\usepackage{algorithmic}
\usepackage{textcomp}
\usepackage{xcolor}
\usepackage{subcaption}
\usepackage{makecell}
\usepackage{amsmath}
\usepackage{hyperref}

\usepackage{graphicx}

\usepackage[T1]{fontenc}
\usepackage{enumitem}
\usepackage{multirow}
\usepackage{tabularx}
\usepackage{graphicx}
\usepackage{romannum}

\DeclareMathOperator*{\argmin}{arg\,min}

\def\BibTeX{{\rm B\kern-.05em{\sc i\kern-.025em b}\kern-.08em
    T\kern-.1667em\lower.7ex\hbox{E}\kern-.125emX}}
    
\makeatletter
\newcommand*{\rom}[1]{\expandafter\@slowromancap\romannumeral #1@}
\makeatother

\begin{document}

\title{ \LARGE \bf Validity Learning on Failures: Mitigating the Distribution Shift in Autonomous Vehicle Planning
}

\author{
Fazel Arasteh, Mohammed Elmahgiubi, Behzad Khamidehi, \\ Hamidreza Mirkhani, Weize Zhang, Cao Tongtong, and Kasra Rezaee\\
{\textit{Noah's Ark Lab, Huawei Technologies Canada}} \\
{\textit{Emails: firstname.lastname@huawei.com}} \\%
}

\maketitle
\begin{abstract}
The planning problem constitutes a fundamental aspect of the autonomous driving framework. Recent strides in representation learning have empowered vehicles to comprehend their surrounding environments, thereby facilitating the integration of learning-based planning strategies. Among these approaches, \textit{Imitation Learning} stands out due to its notable training efficiency. However, traditional \textit{Imitation Learning} methodologies encounter challenges associated with the \textit{co-variate shift} phenomenon. We propose \textit{Validity Learning on Failures, VL(on failure),} as a remedy to address this issue. The essence of our method lies in deploying a pre-trained planner across diverse scenarios. Instances where the planner deviates from its immediate objectives, such as maintaining a safe distance from obstacles or adhering to traffic rules, are flagged as failures. The states corresponding to these failures are compiled into a new dataset, termed the failure dataset. Notably, the absence of expert annotations for this data precludes the applicability of standard imitation learning approaches. To facilitate learning from the closed-loop mistakes, we introduce the \textit{VL} objective which aims to discern valid trajectories within the current environmental context. Experimental evaluations conducted on both reactive CARLA simulation and non-reactive log-replay simulations reveal substantial enhancements in closed-loop metrics such as \textit{Score}, \textit{Progress}, and \textit{Success Rate}, underscoring the effectiveness of the proposed methodology. Further evaluations against the Bench2Drive benchmark demonstrate that \textit{VL(on failure)} outperforms the state-of-the-art methods by a large margin.
\end{abstract}

\section{Introduction}

Autonomous driving is an expanding area within artificial intelligence that has the potential to transform transportation significantly \cite{paden2016survey, chen2023survey2, duarte2018av_impact, AdaptiveNav2022Network}. Planning is a crucial component of autonomous driving systems, involving the generation of safe and efficient trajectories based. We categorize existing approaches into two main groups. Rule-based Optimization Planning, relies on optimization techniques and finite state machine systems for decision-making. These methods use predefined costs and rewards to represent desirable driving behavior, then apply optimization techniques to select the best solution \cite{IDM,ma2015efficient,zhang2022spatial,zhang2022sufficient}. However, they face scalability challenges due to the long-tail problem, which involves handling corner cases and rare events. This issue is particularly problematic in complex urban driving environments, where defining cost functions becomes difficult. While rule-based approaches have been dominant in many industrial applications, creating a robust and comprehensive rule-based planner demands substantial human engineering efforts \cite{apollo2018}. Meanwhile, learning based planning approaches have the potential for scalability by leveraging the vastly available expert human driver's data \cite{gao2020vectornet,TNT,varadarajan2022multipath++}. The availability of large traffic datasets and advances in representation learning has resulted in the vehicles’ ability to understand and predict their surrounding environment paving the road for learning based planning \cite{hu2023uniad,jiang2023vad,li2022bevformer}.
\begin{figure}[t]
    \centering
    \begin{subfigure}[t]{0.4\linewidth}
        \includegraphics[width=1\textwidth]{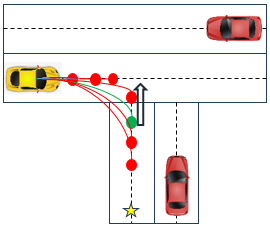}
        \label{fig:ilVvl1}
    \end{subfigure} \qquad
    \begin{subfigure}[t]{0.39\linewidth}
        \includegraphics[width=1\textwidth]{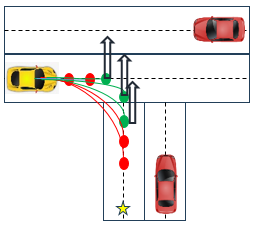}
        \label{fig:ilVvl2}    
    \end{subfigure}
    \caption{\small (Left) \textit{Imitation Learning (IL)}: increase the probability of the candidate trajectory closest to the expert's trajectory (green), (Right) \textit{Validity Learning (VL)}: increase the probability of the valid candidate trajectories (green) }
    \label{fig:IL_VL}
\end{figure}

\noindent Amidst diverse learning-based methodologies, \textit{Imitation Learning (IL)} emerges as a promising avenue owing to its efficacy in training processes. \textit{IL} was first employed as early as 1989 for rural road navigation \cite{26pomerleau1991efficient}. Nonetheless, the basic \textit{IL} suffers from co-variate shift \cite{24ross2011reduction} hindering its generalization beyond the training data. The renowned DAgger \cite{24ross2011reduction} method relies on expert human annotators to create the label, thus making it unviable in situations where such resources are unavailable. \cite{dart} propose an off-policy strategy that introduces noise into the supervisor’s policy during demonstrations. Similarly, \cite{bansal2019chauffeurnet} perturbs the vehicle's current pose uniformly and devises a new smooth trajectory to realign it with the original target position. Yet, rule-based trajectory augmentation methods struggle to accurately represent the motion distribution induced by the learner’s policy, potentially resulting in perturbed driving tendencies. \cite{diffleMPC,urban_driver} utilize a differentiable environment model, enabling actions to receive gradients from multiple future time steps and penalizing actions with significant future divergences. However, these methods' effectiveness hinges on the fidelity of their simulator models. Adversarial \textit{IL} has also been proposed to address this issue \cite{31baram2017end,32ho2016generative,33bhattacharyya2020modeling}; however, as far as we are aware, its application in autonomous vehicle planning remains limited.

\noindent In recent years, there has been a growing focus on enhancing autonomous driving through the combination of imitation learning and reinforcement learning. \cite{bronstein2022hierarchical} introduced a hierarchical model-based imitation learning approach, designed for efficient planning in complex driving tasks. However, the increased computational complexity in real-time applications remains a challenge. \cite{zhang2023learning} tackled the challenge of modeling realistic traffic agents in closed-loop simulations to better test autonomous systems, but the accuracy of their results depends on the quality of the simulated environment. \cite{liu2023blending} blended imitation and reinforcement learning for robust policy improvement, though their approach faces limitations related to the extensive hyperparameter tuning required for scalability across different driving scenarios. Finally, \cite{lu2023imitation} demonstrated that while imitation learning alone is insufficient in difficult driving scenarios, augmenting it with reinforcement learning improves robustness. Yet, this combination requires significant training time, slowing practical deployment.

\noindent In this paper, we introduce \textit{Validity Learning} on failure samples, \textit{VL (on failure)}, as a solution aimed at mitigating the distribution shift problem. The core principle of \textit{VL (on failure)} involves the unrolling of a pre-trained planner across a spectrum of scenarios. Instances where the planner deviates from its immediate objectives, such as maintaining safe distances from obstacles or adhering to traffic regulations, are identified as \textit{failures}. The environments associated with these \textit{failures} are aggregated into a novel dataset referred to as \textit{failure dataset}. Notably, the lack of expert annotations for the \textit{failure dataset} renders conventional \textit{IL} approaches impractical. To alleviate this limitation, we introduce \textit{VL}, a new learning objective to discern valid trajectories within the current environmental context. Fig.\ref{fig:IL_VL} compares \textit{VL} against \textit{IL}. In \textit{IL} (Fig.\ref{fig:IL_VL} (Left)), we train the planner by predicting which candidate trajectory is closest to the expert's action. In \textit{VL} (Fig.\ref{fig:IL_VL} (Right)), however, we train the planner to maximize the probability of the candidate trajectories which are \textit{valid}. A trajectory is \textit{valid} if it conforms to all the safety and traffic rules constraints.  Unlike \textit{IL}, \textit{VL} is a weakly-supervised method that does not need expert annotations. Leveraging \textit{VL}, we directly fine-tune the model using the gathered closed-loop data, obviating the requirement for human labelers.

\noindent We can list the contributions of this paper as the following:
\begin{enumerate}[label=\Roman*.]
    \item We present a systematic approach to accumulate a \textit{failure dataset} derived from planner \textit{mistakes}. The failure dataset consists of data that is either out-of-distribution or under-represented in the expert-labeled dataset.
    \item We present a new learning objective, \textit{Validity Learning (VL)}, aimed at distinguishing \textit{valid} trajectories from \textit{invalid} ones for directly fine-tuning on the failure data, mitigating the distribution shift problem. \textit{VL} does not need human annotation.
    \item We use the reactive CARLA simulation for our data collection. On the other hand, we use a non-reactive log-replay simulation for collecting our \textit{failure dataset}. We present a method for processing the recorded logs that improves consistency between non-reactive and reactive simulation performance. We study this consistency by comparing the closed-loop metrics in reactive and non-reactive simulations.
    \item We compare the performance of \textit{VL} and \textit{IL+RL}, a closely related baseline that combines RL with IL to solve the distribution shift problem. We discuss the benefits and limitations of \textit{VL} versus \textit{IL+RL}. 
    \item We compare the performance of \textit{VL} against the Bench2Drive benchmark. Our \textit{VL} outperforms all the baselines in closed-loop metrics by a large gap.
    
\end{enumerate}
\begin{figure*}[t]
    \centering
    \includegraphics[trim=0 165 385 0, clip, width=\linewidth]{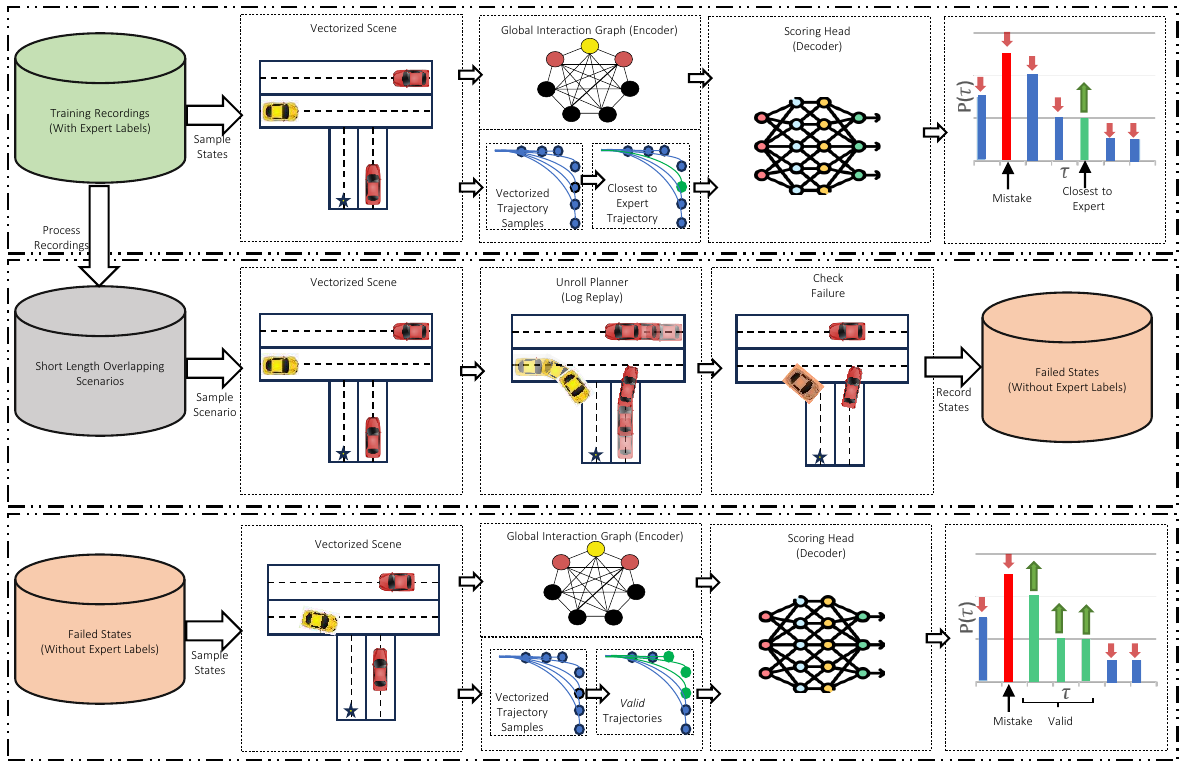}
    \vspace{-0.7cm}
    \caption{\small (Top): \textit{Imitation Learning (IL)} on Expert Labeled States, (Middle): Failure States Data Collection, (Bottom): \textit{Validity Learning (VL)} on Unlabeled Failure States}
    \vspace{-0.5cm}
    \label{fig:IL}
\end{figure*}
\section{Methodology}
\subsection{ Imitation Learning (IL) with Sample-based Planner }
\label{M1}
\noindent Sample-based path planning is a two-staged technique, in which a finite set of fixed length candidate trajectories $ C_{traj}={\tau_1,...,\tau_M}$ is generated first. The candidates are then evaluated to pick the best trajectory that satisfies all the constraints. Fig.\ref{fig:IL}(Top) shows the steps for Imitation Learning of a sample-based planner on the expert dataset. In every scene, we generate a comprehensive set of candidate trajectories toward the ego's desired route (mission) that covers various kinematic behaviors. We use a vectorized representation of the scene similar to \cite{gao2020vectornet}. We use a Transformer Encoder architecture to encode the vectorized scene. The scene embedding is then concatenated with the vectorized representation of each trajectory sample and passed to a fully connected scoring head. The scoring head assigns each trajectory a score logit. We take a soft-max over the logits to get the probability distribution over the trajectory samples ($P(\tau)$). We use Imitation Learning with cross-entropy loss (eq.\ref{eq:imitation_loss}) to train the planner:
\begin{equation}
\begin{split}
    l_{imitation} = - log(P(\tau))  \qquad \\ 
    \tau= \argmin( dist(\tau_c,\tau_{expert}) ), \tau_c\in C_{traj}, 
    \label{eq:imitation_loss} 
\end{split}
\end{equation}
where $dist(\tau_1, \tau_2) = \lVert \tau_1 - \tau_2\rVert^2$.
\subsection{Failure States Data Collection}

\noindent Our experiments with the sample-based planner trained using imitation learning (IL) confirm that strong open-loop performance metrics (e.g. Final Displacement Error (FDE)) do not necessarily translate to robust closed-loop performance outcomes (e.g., Collision, Success) \cite{leader_board1,jia2024bench}. We attribute this discrepancy primarily to an unavoidable distribution shift problem. Essentially, once the model is deployed in a sequential decision-making process, it encounters out-of-distribution states, leading to catastrophic failures. Consequently, our focus shifts to identifying and learning from these failure states by analyzing the planner's errors in these scenarios.  

\noindent Fig\ref{fig:IL}(Middle)  outlines the process of collecting data where the planner fails. Before data collection, we first segment each recorded log into fixed-length overlapping short scenarios, initiated at regular intervals. For example, if an expert's recording lasts for 60 seconds, the first scenario spans from 0 to 12 seconds, the second from 3 to 15 seconds, and the last scenario from 48 to 60. This segmentation method was adopted to limit the effect of the deviations between the ego and the expert location during the log-replay simulation. Notably, significant deviation from the expert's path could render the recorded logs useless for training. For instance, in a Parking Exit Scenario in CARLA, the adversary agent waits for the ego vehicle to reach a specific point before initiating its exit. However, this behavior of the adversary agent is not enforceable in the log-replay simulation. By splitting the logs in this manner, we ensure that the ego vehicle encounters the same adversary scenario consistently during the non-reactive data replay simulations.

\noindent For each training scenario, we unroll the planner's actions for the ego vehicle while other agents follow a log replay, continuing until the ego vehicle either reaches its goal or fails. A \textit{planner mistake} occurs when the planner incorrectly selects an \textit{invalid} trajectory as its action. A trajectory $\tau$ is deemed \textit{invalid} in a given environment if it meets any of the following conditions otherwise, the trajectory is deemed \textit{valid}:
\begin{itemize}
    \item Veers off the road
    \item Collides with other traffic agents
    \item Violates safety protocols
    \item Disregards traffic rules (e.g., running a red light)
    \item Compromises comfort
    \item Hinders progress (i.e., the vehicle becomes stuck and stops moving)
\end{itemize}

\noindent In the event of a planner failure, we collect all state observations from the scenario and compile them into a new dataset known as \textit{Failure States}. Note that the expert's action is not know in the \textit{Failure States}.

\subsection{Validity Learning (VL) on Failure States}

\noindent When a \textit{planner mistake} occurs, the model assigns a high probability to an \textit{invalid} trajectory. The objective is to fine-tune the model using \textit{Validity Learning (VL)}, which essentially involves reducing the probability of \textit{invalid} trajectories while increasing the probability of \textit{valid} ones. Fig\ref{fig:IL}(Bottom) illustrates the steps involved in \textit{Validity Learning}. Similar to \textit{Imitation Learning}, for each scenario, we generate a set of $M$ candidate trajectories, denoted as $C_{traj}=\{\tau_1,...,\tau_M\}$, and then evaluate these candidates. Given the assumption that other traffic agents are non-reactive to the ego's actions and will adhere to their recorded logs, we can infer the future state of the environment at any given time. Consequently, it becomes straightforward to verify the validity of a trajectory $\tau$ by comparing it against the known future state of the environment. For instance, we can assess whether $\tau$ will result in a collision within the planning horizon by checking it against the known future locations of all other agents. We define $C_{valid}=\{ \tau \in C_{traj} | \tau \text{ is \textit{valid}} \}$ as the set of all valid candidate trajectories. The planner is fine-tuned by maximizing the approximated probability of the $C_{valid}$, which is equivalent to minimizing the $l_{valid}$ (eq.\ref{eq:weak_sup_loss}). Note that, by definition, a \textit{planner mistake} invariably leads to a high $l_{valid}$, which provides a substantial training signal.

\noindent To prevent forgetting the policy learned during the \textit{IL} phase, we continue training on expert data as well. Specifically, during each training step, we select one mini-batch of failure data and one mini-batch of expert data. The total loss is calculated by weighted sum of the \textit{validity loss} (eq.\ref{eq:weak_sup_loss}) on the failure mini-batch and the \textit{imitation loss} (eq.\ref{eq:imitation_loss}) on the expert mini-batch, as shown in eq.\ref{eq:total_loss}. 

\noindent Following one epoch of \textit{Validity Learning}, we can return to data collection to gather additional failure data by unrolling the newly fine-tuned model. This approach allows for incremental performance improvements.
\begin{equation}
\begin{split}
    l_{valid} = -\log \left(\sum{P(\tau)} \right), \tau \in C_{valid} \\
    C_{valid}=\{ \tau \in C_{traj} | \tau \text{ is \textit{valid}} \}
    \label{eq:weak_sup_loss}   
\end{split}
\end{equation}

\begin{equation}
l = w_{valid}.l_{valid} + w_{imitation}.l_{imitation} \label{eq:total_loss}
\end{equation}

\section{Evaluation}
\subsection{Experiments Setup}
\noindent\textbf{Simulation Environment:} We use  CARLA Leaderboard 2 for all our experiments. CARLA \cite{carla} is an open-source simulator for autonomous driving research providing a highly realistic environment to test and develop autonomous vehicles under diverse urban settings, dynamic weather conditions, and various traffic scenarios. CARLA allows researchers to design and train AI models for tasks such as perception, decision-making, and control in a controlled yet highly detailed virtual world. CARLA Leaderboard 2 extends the original CARLA Leaderboard by offering a set of standardized tasks and metrics for assessing the performance of autonomous driving systems. The leaderboard features a variety of scenarios and challenges that test different aspects of driving.

\noindent\textbf{Data Collection:}
A significant challenge in utilizing the CARLA Leaderboard 2 lies in the absence of a single expert capable of consistently performing across the diverse range of scenarios presented in the benchmark. This limitation arises due to the wide spectrum of driving conditions, from intricate urban environments to complex traffic situations, that the benchmark encompasses. As a result, high-quality expert data demonstrating optimal driving behaviors across all scenarios is lacking. This deficiency complicates both the training and evaluation of autonomous driving systems, as the absence of comprehensive expert demonstrations hinders the development of robust and generalized driving policies. Additionally, the original CARLA training scenarios feature extended routes, which diminishes their utility for data collection. Longer routes are more difficult to complete successfully, as the probability of failure increases with route length. To address these limitations, we curated a balanced set of CARLA short-segmented scenarios, inspired by Bench2Drive \cite{jia2024bench}, for data collection purposes. A rule-based classical planner, fine-tuned manually with privileged perception data from CARLA, was employed to generate our expert dataset \cite{slt,slt1}. Each recorded log was further segmented into 12 seconds long scenarios seperated with a gap of 2.5 seconds to create the training scenarios for the data collection phase.

\begin{table}[t]
\centering
\label{table_trained_constraint}
\caption{ CARLA reactive evaluation over the 220 short segment scenarios of Bench2Drive}
\begin{tabularx}{0.5 \textwidth}{>{\raggedright\arraybackslash}m{1.1cm}|X|X|X}
\noalign{\hrule height 0.05cm}
\textbf{Input} & \textbf{Method} & \textbf{DrivingScore$\uparrow$ } & \textbf{SuccessRate(\%)$\uparrow$} \\
\hline
\multirow{4}{1cm}{Privileged} 
& IL & 59.96 & 48.69 \\
& VL (on expert) & 60.01 & 49.23 \\ 
& IL + RL & 74.57 & 60.41 \\
& VL (on failure) & \textbf{77.30} & \textbf{72.25} \\ 
& Expert & 82.37 & 84.29  \\
\hline
\multirow{10}{8cm}{Camera} 
& AD-MLP\cite{zhai2023rethinking} & 18.05 & 00.00 \\ 
& UniAD-Base\cite{hu2023planning} & 45.81 & 16.36 \\ 
& VAD\cite{jiang2023vad} & 42.35 & 15.00 \\ 
& TCP-traj\cite{wu2022trajectory} & 59.90 & 30.00 \\ 
& ThinkTwice\cite{jia2023think} & 62.44 & 31.23 \\ 
& DriveAdapter\cite{jia2023driveadapter} & 64.22 & 33.08 \\ 
& VL (on failure) & \textbf{73.29} & \textbf{65.44} \\ 
& Expert & 75.82 & 82.72  \\
\noalign{\hrule height 0.05cm}
\end{tabularx}
\label{tab:grouped_results}
\end{table}

\begin{table}[t]
\centering
\label{tab:non-reactive}
\caption{ Non-reactive closed-loop evaluation over the test-set recording logs}
\begin{tabularx}{0.5 \textwidth}{>{\raggedright\arraybackslash}m{1.75cm}|X|X|X|X}
\noalign{\hrule height 0.05cm}
\textbf{Method} & \textbf{Progress(\%)$\uparrow$ } & \textbf{Success(\%)$\uparrow$} & \textbf{Collision(\%)$\downarrow$} & \textbf{MDBC\textit{(m)}$\uparrow$} \\
\hline
IL & 79.75 & 69.00 & 29.08 & 172.39 \\
VL (on expert) & 81.24 & 70.20 & 28.14 & 178.44  \\ 
IL + RL & 83.62 & 75.15 & 22.08 & 235.95\\
VL (on failure) & \textbf{87.09} & \textbf{82.16} & \textbf{15.71} & \textbf{344.50} \\ 
\noalign{\hrule height 0.05cm}
\end{tabularx}
\label{tab:non-reactive}
\end{table}
\noindent\textbf{Evaluation Benchmark:} We use the Bench2Drive \cite{jia2024bench} for our bench-marking. Bench2Drive provides a comprehensive evaluation scenario set consisting of 220 short-segment scenarios that cover various challenging traffic behaviors. These short segments are particularly valuable because, on long routes, it becomes difficult to distinguish the performance of different algorithms, as the complexity of long routes is too challenging for most algorithms.

\noindent Furthermore, we also evaluate our models in the non-reactive log-replay simulation. For this purpose, we curated a separate balanced set of CARLA short-segmented scenarios to collect our test recordings. We process the test recordings into short length overlapping scenarios.

\noindent\textbf{Evaluation Metrics:} We focus on two key metrics reported in Bench2Drive: 
\begin{enumerate}[label=\Roman*.]
    \item Driving Score: driving score evaluates the quality of driving behavior by penalizing infractions such as collisions, off-road driving, or failure to follow traffic rules as defined in the official CARLA Leaderboard2 \cite{carla}.
    \item Success Rate: success rate measures the percentage of scenarios where the vehicle successfully reaches its goal without encountering critical failures, defined as any failure except a 'min-speed-infraction', where the vehicle momentarily drops below a certain speed threshold.
\end{enumerate}

\noindent For non-reactive evaluation, we report the following metrics:

\begin{enumerate}[label=\Roman*.]
    \item Progress\%: progress rate expresses the average ratio of the distance driven by the vehicle towards the goal compared to the ground truth before encountering any failures. We terminate a scenario if the ego makes a \textit{mistake} (collision, veering off the road, etc)
    \item Success\%: success rate denotes the percent of scenarios where the ego successfully reaches its destination without any failures (progress is 100\%).
    \item Collision\%: collision rate represents the percent of scenarios that the ego at least encounters one collision.
    \item MDBC: mean Distance between collision represents the average driving distance (meters) of the ego before getting into a collisions. 
\end{enumerate}

\noindent\textbf{Baselines:} Other than the baselines reported in Bench2Drive, we study the performance the following baselines:\\
\begin{enumerate}[label=\Roman*.]
\item Expert: this baseline reports the performance of our expert classical planner that we have fine-tuned for solving the CARLA scenarios to collect our expert dataset.
\item Imitation Learning (IL): we train our planner on the expert dataset with only the imitation loss (eq.\ref{eq:imitation_loss}).
\item  Combined imitation and reinforcement learning \textit{(IL+RL)} \cite{lu2023imitation}: this is a closely related baseline that combines imitation learning with reinforcement learning (RL). This hybrid approach supplements IL with RL to enhance decision-making in rare or challenging events. Since the action space of our planner is discrete, we adapted our pipeline to implement a q-learning version of \cite{lu2023imitation}. Specifically, we treated the scoring head as the Q-network and we trained it with the TD-loss (eq.\ref{eq:td}) for RL transitions and imitation loss (eq.\ref{eq:imitation_loss}) for expert data. Everything else remains identical to \cite{lu2023imitation}.

\begin{equation}
    \delta_t = R(s_t, \tau_t) + \gamma \max_{\tau \in C_{traj}} Q(s_{t+1}, \tau) - Q(s_t, \tau_t)
    \label{eq:td}
\end{equation}
\end{enumerate}

\noindent\textbf{Training:} We first train the model with \textit{IL} for 125K training steps with a batch size of 64. We start collecting failure states by unrolling 4\% of the training scenarios at every epoch. In the \textit{VL} phase, we use a batch size of 64 for the expert data and 32 for the failure samples. We continue training for 200K more training steps. To keep the baselines comparable, we train the \textit{IL} baseline for 350K steps. In the case of \textit{IL+RL}, we first train the \textit{IL} for 125K steps and continue training with \textit{RL} for 375K more training steps.

\begin{figure}[t]
    \centering
    \includegraphics[trim=0 300 520 0, clip, width=\linewidth]{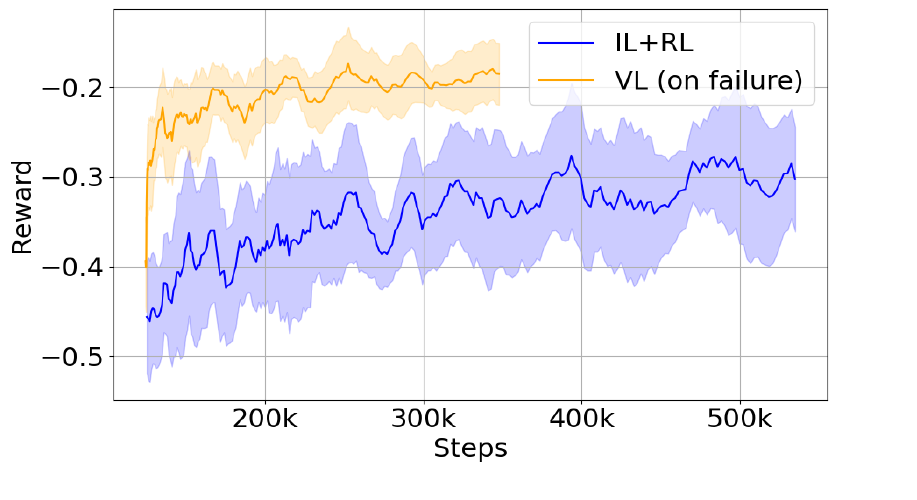}
    \caption{\small VL(on failure) vs IL+RL: Reward over Training Scenarios (Moving Average, and 95\% confidence intervals) }
    \label{fig:vlVSrl}
\end{figure}
\begin{figure}[t]
    \centering
    \includegraphics[trim=0 300 535 0, clip, width=0.92\linewidth]{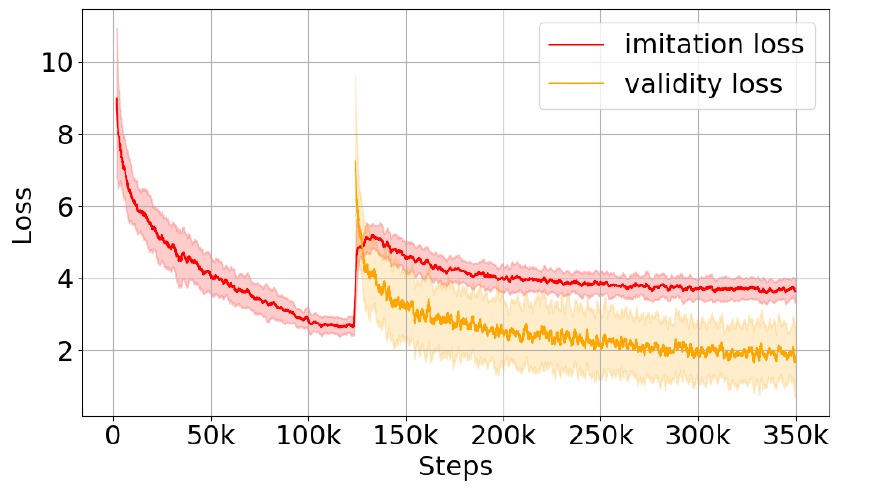}
    \caption{\small Effect of validity loss on imitation loss}
    \label{fig:validity_loss_effect}
\end{figure}

\subsection{Experimental Results}

\noindent We conduct our evaluation on CARLA with privileged input information. We create the vectorized scene representation by directly querying CARLA for the exact information about the scene. TABLE.\ref{tab:grouped_results} reports the closed-loop performance of various methods over the 220 short segment scenarios of Bench2Drive \cite{jia2024bench}. Accordingly, validity learning on failure samples, \textit{VL (on failure)}, outperforms \textit{IL} and \textit{IL+RL} baselines in both \textit{Driving Score}, and \textit{Success Rate}, approaching the performance of our data collection Expert. Fig.\ref{fig:vlVSrl} compares the reward over the training scenarios during the training for \textit{VL (on failure)} and \textit{IL+RL}. Note that the first 125K training steps are imitation learning only. We observe that \textit{VL (on failure)} can quickly converge to high values of average reward, however, IL+RL takes a much longer time to reach inferior average rewards. We believe this is largely due to the sample efficiency of \textit{VL}. As eq.\ref{eq:weak_sup_loss} suggests, validity loss evaluates all candidate actions simultaneously with respect to each other. In contrast, RL suffers from lower sample efficiency, as the TD-loss can evaluate only one state-action pair at a time. As a result, learning an effective RL policy requires much longer training and computation power. 

\noindent Furthermore, to compare the performance of our model with the baselines studied in \cite{jia2024bench}, we evaluate our model with the camera input as well. Specifically, we deployed 6 camera sensors around the car and fine-tuned a version of the VAD backbone to estimate the vectorized scene representation \cite{jiang2023vad, slt}. As TABLE.\ref{tab:grouped_results} suggests, \textit{VL (on failure)} out-performs all the baselines with a large margin.

\subsection{Ablation Study}
  
\noindent To further investigate the significance of incorporating failure data into the training process, we train a planner utilizing validity loss (eq.\ref{eq:weak_sup_loss}), but restrict the training solely to expert data, which notably contains no failure examples. This variant is referred to as \textit{VL (on expert)}. As shown in TABLE.\ref{tab:grouped_results} and TABLE\ref{tab:non-reactive}, the performance of the \textit{VL (on expert)} model does not exhibit substantial improvements over the \textit{IL} baseline. In contrast, training with \textit{VL (on failure)} data leads to a marked enhancement in closed-loop metrics. These findings underscore the critical role of failure data in achieving significant performance gains, suggesting that exposure to suboptimal or failure scenarios is crucial for the model to develop robust decision-making capabilities in real-world conditions.

\noindent Figure Fig\ref{fig:validity_loss_effect} presents the trajectories of imitation loss and validity loss during the course of training. It is important to highlight that validity training is initiated at approximately the 125,000th training step. A pronounced and immediate increase in imitation loss is observed coinciding with the onset of validity training. Despite this spike, the imitation loss does not revert to its pre-validity training levels, even as we observe consistent improvements in closed-loop performance metrics as shown in \ref{fig:vlVSrl}. This discrepancy suggests a complex interaction between the two loss functions, where the reduction in validity loss is contributing to the enhanced closed-loop performance. The inability of imitation loss to fully recover indicates the critical role of validity loss in driving the observed improvements in the overall system performance.

\subsection{Limitations and Discussions}

\noindent It is important to recognize that one of the limitations of \textit{VL} is that it assumes that the environment is non-reactive, meaning that it remains static and does not respond dynamically to the agent's actions. This allows for the evaluation of a trajectory's validity over the entire planning horizon.  In general, where this assumption does not hold, \textit{VL} may not be applicable. However, in autonomous driving, where fully reactive simulation is challenging, the non-reactive simulation can provide a reasonable proxy \cite{NAVSIM}. Our experiments with non-reactive simulations with the raw CARLA recordings shows that non-reactive closed loop metrics are not necessarily correlated with the reactive CARLA metrics. In non-reactive simulation, significant deviation from the expert’s path can render the CARLA adversary scenarios useless especially if the scenario is long. However, processing the raw recordings into the fixed-length overlapping short scenarios as described in Data Collection section can mitigate this problem. TABLE. \ref{tab:non-reactive} reports the non-reactive closed-loop evaluation metrics over the test-set processed recording logs. Comparing TABLE.\ref{tab:grouped_results} and TABLE.\ref{tab:non-reactive}, we can observe that the non-reactive closed loop metrics are correlated with the reactive CARLA metrics. In other words, better non-reactive metrics lead to better CARLA metrics. \textit{VL(on failure)} outperforms \textit{IL} and \textit{IL+RL} in non-reactive simulation metrics.

\noindent Another limitation of \textit{VL} is its assumption that the validity of a trajectory can be determined solely over a fixed planning horizon, without considering the states that follow after the trajectory is executed. For instance, a trajectory may appear valid within the planning horizon (avoiding collisions, staying on the road, etc.), but it might steer too close to a construction site, trapping the vehicle in a situation where it cannot proceed without risking a collision. This issue could be mitigated by extending the planning horizon for generated trajectory samples, as a longer horizon would allow for better foresight into potential future problems. Our empirical results indicate that the selected 3-second planning horizon is sufficiently long to handle many of the CARLA scenarios, although it may not be adequate for all cases.
\section{Conclusion}
In conclusion, in this paper, we have explored the efficacy of Imitation Learning for autonomous vehicle planning and proposed a novel approach, \textit{Validity Learning on Failures, VL (on failures)}, to mitigate the challenges associated with co-variate shift. Leveraging a pre-trained planner and identifying deviations as learning opportunities, \textit{VL} enriches the planning process with closed-loop mistakes data. Though lacking expert annotations, this data becomes invaluable through the \textit{VL} objective to discern valid trajectories within various environmental contexts. Our experimental findings, validated on non-reactive log-replay and reactive CARLA simulations, underscore significant improvements in critical planning metrics such as Driving Score, Progress, and Collision Rate compared to the state-of-the-art schemes. However, limitations such as the reliance on non-reactive simulations and the fixed planning horizon suggest areas for further research. Overall,\textit{ VL on failure} provides a promising direction for more reliable and scalable autonomous driving systems.

\bibliographystyle{ieeetr}
\bibliography{Bib}

\end{document}